\documentclass[11pt,a4paper]{article}

\usepackage[hyperref]{emnlp2018}
\usepackage{times}
\usepackage{latexsym}
\usepackage{url}
\usepackage{graphicx}
\usepackage{subcaption}
\usepackage{enumitem}
\usepackage[normalem]{ulem}
\usepackage{amsmath,amssymb}
\usepackage{dblfloatfix}
\usepackage{float}
\usepackage{multicol}
\usepackage{multirow}
\usepackage{array}
\usepackage{tabu,colortbl}
\usepackage{hhline}
\usepackage[skip=2pt,font=footnotesize]{caption}
\usepackage{textcomp}

\usepackage{tikz}
\usepackage{ragged2e}

\usepackage{esvect}

\definecolor{box_background}{rgb}{0.84,0.92,1}
\colorlet{pale_gray}{Plum!5}
\colorlet{pale_red}{Plum!10}

\newcommand{\CM}{\cellcolor{pale_red}}

\aclfinalcopy

\newcommand{\f}{{\bf f}}
\newcommand{\p}{{$\phantom{0}$}}

\newcommand{\tr}[1][3pt]{\mathrel{%
   \hbox{\rule[\dimexpr\fontdimen22\textfont2-.2pt\relax]{#1}{.4pt}}%
   \mkern-4mu\hbox{\usefont{U}{lasy}{m}{n}\symbol{41}}}}

\title{Contextual Parameter Generation for \\ Universal Neural Machine Translation}

\author{Emmanouil Antonios Platanios$^\dagger$, Mrinmaya Sachan$^\dagger$, Graham Neubig$^\ddag$, Tom M. Mitchell$^\dagger$ \\
        $^\dagger$Machine Learning Department, $^\ddag$Language Technologies Institute \\
        Carnegie Mellon University \\
        \texttt{\{e.a.platanios,mrinmays,gneubig,tom.mitchell\}@cs.cmu.edu}}

\date{}

\begin{document}
\maketitle
\begin{abstract}

We propose a simple modification to existing neural machine translation (NMT) models that enables using a single {\em universal} model to translate between multiple languages while allowing for language specific parameterization, and that can also be used for {\em domain adaptation}. 
Our approach requires no changes to the model architecture of a standard NMT system, but instead introduces a new component, the {\em contextual parameter generator} (CPG), that generates the parameters of the system (e.g., weights in a neural network). 
This parameter generator accepts source and target language embeddings as input, and generates the parameters for the encoder and the decoder, respectively. 
The rest of the model remains unchanged and is shared across all languages. 
We show how this simple modification enables the system to use monolingual data for training and also perform {\em zero-shot} translation. 
We further show it is able to surpass state-of-the-art performance for both the IWSLT-15 and IWSLT-17 datasets and that the learned language embeddings are able to uncover interesting relationships between languages.

\end{abstract}



\section{Introduction}

Neural Machine Translation (NMT) directly models the mapping of a source language to a target language without any need for training or tuning any component of the system separately. 
This has led to a rapid progress in NMT and its successful adoption in many large-scale settings \cite{Wu:2016,Crego:2016}.
The encoder-decoder abstraction makes it conceptually feasible to build a system that maps any source sentence in any language to a vector representation, and then decodes this representation into any target language. 
Thus, various approaches have been proposed to extend this abstraction for multilingual MT \cite{Luong:2015a,Dong:2015,Johnson:2017,Ha:2016,Firat:2016a}.


Prior work in multilingual NMT can be broadly categorized into two paradigms. 
The first, {\em universal NMT} \citep{Johnson:2017,Ha:2016}, uses a single model for all languages. 
Universal NMT lacks any language-specific parameterization, which is an oversimplification and detrimental when we have very different languages and limited training data.
As verified by our experiments, the method of \citet{Johnson:2017} suffers from high sample complexity and thus underperforms in limited data settings.
The universal model proposed by \citet{Ha:2016} requires a new coding scheme for the input sentences, which results in large vocabulary sizes that are difficult to scale.
The second paradigm, {\em per-language encoder-decoder} \citep{Luong:2015a,Firat:2016a}, uses separate encoders and decoders for each language. 
This does not allow for sharing of information across languages, which can result in overparameterization and can be detrimental when the languages are similar.

In this paper, we strike a balance between these two approaches, proposing a model that has the ability to learn parameters separately for each language, but also share information between similar languages. 
We propose using a new \textit{contextual parameter generator} (CPG) which (a) generalizes all of these methods, and (b) mitigates the aforementioned issues of \emph{universal} and \emph{per-language encoder-decoder} systems.
It learns language embeddings as a context for translation and uses them to generate the parameters of a shared translation model for all language pairs.
Thus, it provides these models the ability to learn parameters separately for each language, but also share information between similar languages. 
The parameter generator is general and allows any existing NMT model to be enhanced in this way.\footnote{In fact, it could likely be applied in other scenarios, such as domain adaptation, as well.} 
In addition, it has the following desirable features:
\begin{enumerate}[noitemsep,leftmargin=*,topsep=2pt]
	\item \textbf{Simple:} Similar to \citet{Johnson:2017} and \citet{Ha:2016}, and in contrast with \citet{Luong:2015a} and \citet{Firat:2016a}, it can be applied to most existing NMT systems with some minor modification, and it is able to accommodate attention layers seamlessly.
	\item \textbf{Multilingual:} Enables multilingual translation using the same single model as before.
	\item \textbf{Semi-supervised:} Can use monolingual data.
	\item \textbf{Scalable:} Reduces the number of parameters by employing extensive, yet controllable, sharing across languages, thus mitigating the need for large amounts of data, as in \citet{Johnson:2017}. It also allows for the decoupling of languages, avoiding the need for a large shared vocabulary, as in \citet{Ha:2016}.
	\item \textbf{Adaptable:} Can adapt to support new languages, without requiring complete retraining.
	\item \textbf{State-of-the-art:} Achieves better performance than pairwise NMT models and \citet{Johnson:2017}. In fact, our approach can surpass state-of-the-art performance.
\end{enumerate}
We first introduce a modular framework that can be used to define and describe most existing NMT systems. 
Then, in Section \ref{sec:proposed_method}, we introduce our main contribution, the \textit{contextual parameter generator} (CPG), in terms of that framework.
We also argue that the proposed approach takes us a step closer to a common universal interlingua.

\section{Background}
\label{sec:background}

We first define the multi-lingual NMT setting and then introduce a modular framework that can be used to define and describe most existing NMT systems. 
This will help us distill previous contributions and introduce ours.

\paragraph{Setting.} We assume that we have a set of source languages $S$ and a set of target languages $T$. 
The total number of languages is $L = |S \cup T|$. 
We also assume we have a set of $C \leq |S| \times |T|$ pairwise parallel corpora, $\{P_1, \dots, P_C\}$, each of which contains a set of sentence pairs for a single source-target language combination. 
The goal of multilingual NMT is to build a model that, when trained using the provided parallel corpora, can learn to translate well between any pair of languages in $S \times T$. 
The majority of related work only considers pairwise NMT, where $|S|=|T|=1$.

\subsection{NMT Modules}
\label{sec:nmt_modules}

\begin{figure*}[!t]
    \centering
    \includegraphics[width=1.0\textwidth,clip,trim=0mm 0mm 0mm 2mm]{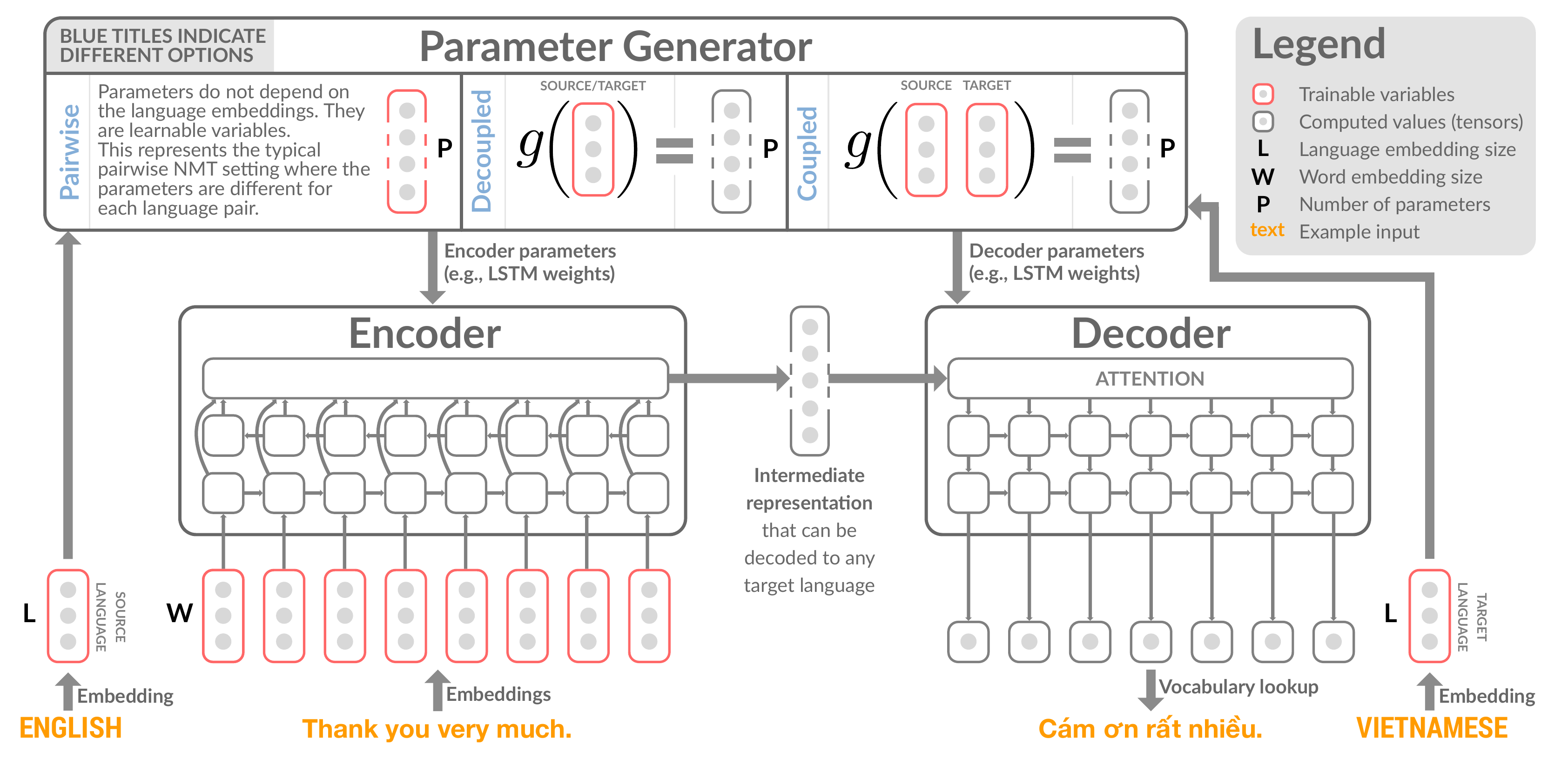}
	\vspace{-1.3em}
	\caption{Overview of an NMT system, under our modular framework. Our main contribution lies in the parameter generator module (i.e., coupled or decoupled --- each of the boxes with blue titles is a separate option). Note that $g$ denotes a parameter generator network. In our experiments, we consider linear forms for this network. However, our contribution does not depend on the choices made regarding the rest of the modules; we could still use our parameter generator with different architectures for the encoder and the decoder, as well as using different kinds of vocabularies.}
	\label{fig:modular_nmt}
	\vspace{-1.3em}
\end{figure*}

Most NMT systems can be decomposed to the following modules (also visualized in Figure \ref{fig:modular_nmt}).

\paragraph{Preprocessing Pipeline.} The data preprocessing pipeline handles tokenization, cleaning, normalizing the text data and building a {\em vocabulary}, i.e. a two-way mapping from preprocessed sentences to sequences of word indices that will be used for the translation. 
A commonly used proposal for defining the vocabulary is the byte-pair encoding (BPE) algorithm which generates subword unit vocabularies \citep{Sennrich:2016a}. 
This eliminates the notion of out-of-vocabulary words, often resulting in increased translation quality.

\paragraph{Encoder/Decoder.} The {\em encoder} takes in indexed source language sentences, and produces an intermediate representation that can later be used by a {\em decoder} to generate sentences in a  target language. 
Generally, we can think of the encoder as a function, $\f^{(enc)}$, parameterized by ${\bf \theta}^{(enc)}$. 
Similarly, we can think of the decoder as another function, $\f^{(dec)}$, parameterized by ${\bf \theta}^{(dec)}$. 
The goal of learning to translate can then be defined as finding the values for ${\bf \theta}^{(enc)}$ and ${\bf \theta}^{(dec)}$ that result in the best translations. 
A large amount of previous work proposes novel designs for the encoder/decoder module.
For example, using attention over the input sequence while decoding \cite{Bahdanau:2014,Luong:2015b} provides significant gains in translation performance.\footnote{Note that depending on the vocabulary that is used and on whether it is one shared vocabulary across all languages, or one vocabulary per language, the output projection layer of the decoder (which produces probabilities over words) may be language dependent, or common across all languages. In our experiments, we used separate vocabularies and thus this layer was language-dependent.}

\paragraph{Parameter Generator.} All modules defined so far have previously been used when describing NMT systems and are thus easy to conceptualize. 
However, in previous work, most models are trained for a given language pair, and it is not trivial to extend them to work for multiple pairs of languages. 
We introduce here the concept of the {\em parameter generator}, which makes it easy to define and describe multilingual NMT systems. 
This module is responsible for generating ${\bf \theta}^{(enc)}$ and ${\bf \theta}^{(dec)}$ for any given source and target language. 
Different parameter generators result in different numbers of learnable parameters and can thus be used to share information across different languages. 
Next, we describe related work, in terms of the parameter generator for NMT:
\begin{itemize}[noitemsep,leftmargin=*]
	\item \textbf{Pairwise:} In the simple and commonly used pairwise NMT setting \cite{Wu:2016,Crego:2016}, the parameter generator would generate separate parameters, ${\bf \theta}^{(enc)}$ and ${\bf \theta}^{(dec)}$, for each pair of source-target languages. This results in no parameter sharing across languages, and thus $\mathcal{O}(ST)$ parameters.
	\item \textbf{Per-Language:} In the case of \citet{Dong:2015}, \citet{Luong:2015a} and \citet{Firat:2016a}, the parameter generator would generate separate encoder parameters, ${\bf \theta}^{(enc)}$, for each source language, and separate decoder parameters, ${\bf \theta}^{(dec)}$, for each target language. This leads to a reduction in the number of learnable parameters for multilingual NMT, from $\mathcal{O}(ST)$ to $\mathcal{O}(S+T)$. On one hand, \citet{Dong:2015} train multiple models as a one-to-many multilingual NMT system that translates from one source language to multiple target languages. On the other hand, \citet{Luong:2015a} and \citet{Firat:2016a} perform many-to-many translation. \citet{Luong:2015a}, however, only report results for a single language pair and do not attempt multilingual translation. \citet{Firat:2016a} propose an attention mechanism that is shared across all language pairs. We generalize the idea of multi-way multilingual NMT with the parameter generator network, described later.
	\item \textbf{Universal:} In the case of \citet{Ha:2016} and \citet{Johnson:2017}, the authors propose using a single common set of encoder-decoder parameters for all language pairs. While \citet{Ha:2016} embed words in a common semantic space across languages, \citet{Johnson:2017} learn language embeddings that are in the same space as the word embeddings. Here, the parameter generator would provide the same parameters ${\bf \theta}^{(enc)}$ and ${\bf \theta}^{(dec)}$ for all language pairs. It would also create and keep track of learnable variables representing language embeddings that are prepended to the encoder input sequence. As we observed in our experiments, this system fails to perform well when the training data is limited. Finally, we believe that embedding languages in the same space as words is not intuitive; in our approach, languages are embedded in a separate space.
\end{itemize}

In contrast to all these related systems, we provide a simple, efficient, yet effective alternative -- a parameter generator for multilingual NMT, that enables semi-supervised and zero-shot learning. 
We also learn language embeddings, similar to \citet{Johnson:2017}, but in our case they are separate from the word embeddings and are treated as a {\em context} for the translation, in a sense that will become clear in the next section. 
This notion of  {\em context} is used to define parameter sharing across various encoders and decoders, and, as we discuss in our conclusion, is even applicable beyond NMT.

\section{Proposed Method}
\label{sec:proposed_method}

We propose a new way to share information across different languages and to control the amount of sharing, through the parameter generator module.
More specifically, we propose {\em contextual parameter generators}.

\paragraph{Contextual Parameter Generator.} Let us denote the source language for a given sentence pair by $\ell_s$ and the target language by $\ell_t$. 
Then, when using the contextual parameter generator, the parameters of the encoder are defined as $\theta^{(enc)}\triangleq g^{(enc)}(\bf{l}_s)$, for some function $g^{(enc)}$, where $\bf{l}_s$ denotes a language embedding for the source language $\ell_s$. 
Similarly, the parameters of the decoder are defined as $\theta^{(dec)}\triangleq g^{(dec)}(\bf{l}_t)$ for some function $g^{(dec)}$, where $\bf{l}_t$ denotes a language embedding for the target language $\ell_t$. 
Our generic formulation does not impose any constraints on the functional form of $g^{(enc)}$ and $g^{(dec)}$. 
In this case, we can think of the source language, $\ell_s$, as a context for the encoder. 
The parameters of the encoder depend on its context, but its architecture is common across all contexts. 
We can make a similar argument for the decoder, and that is where the name of this parameter generator comes from. 
We can even go a step further and have a parameter generator that defines $\theta^{(enc)}\triangleq g^{(enc)}(\bf{l}_s, \bf{l}_t)$ and $\theta^{(dec)}\triangleq g^{(dec)}(\bf{l}_s, \bf{l}_t)$, thus coupling the encoding and decoding stages for a given language pair.
In our experiments we stick to the previous, {\em decoupled}, form, because unlike \citet{Johnson:2017}, it has the potential to lead to an {\em interlingua}.

Concretely, because the encoding and decoding stages are decoupled, the encoder is not aware of the target language while generating it.
Thus, we can take an encoded intermediate representation of a sentence and translate it to any target language. 
This is because, in this case, the intermediate representation is independent of any target language.
This makes for a stronger argument that the intermediate representation produced by our encoder could be approaching a universal interlingua, more so than methods that are aware of the target language when they perform encoding.

\subsection{Parameter Generator Network}
\label{sec:parameter_generator_network}

We refer to the functions $g^{(enc)}$ and $g^{(dec)}$ as {\em parameter generator networks}. 
Even though our proposed NMT framework does not rely on a specific choice for $g^{(enc)}$ and $g^{(dec)}$, here we describe the functional form we used for our experiments. 
Our goal is to provide a simple form that works, and for which we can reason about. 
For this reason, we decided to define the parameter generator networks as simple linear transforms, similar to the factored adaptation model of \citet{Michel:2018}, which was only applied to the bias terms of the output softmax:
\begin{align}
	g^{(enc)}(\bf{l}_s) & \triangleq \bf{W^{(enc)}} \bf{l}_s, \\ 
	g^{(dec)}(\bf{l}_t) & \triangleq \bf{W^{(dec)}} \bf{l}_t,
\end{align}
where $\bf{l}_s, \bf{l}_t \in \mathbb{R}^M$, $\bf{W^{(enc)}} \in \mathbb{R}^{P^{(enc)} \times M}$, $\bf{W^{(dec)}} \in \mathbb{R}^{P^{(dec)} \times M}$, $M$ is the language embedding size, $P^{(enc)}$ is the number of parameters of the encoder, and $P^{(dec)}$ is the number of parameters of the decoder.

Another way to interpret this model is that it imposes a low-rank constraint on the parameters.
As opposed to our approach, in the base case of using multiple pairwise models to perform multilingual translation, each model has $P = P^{(enc)} + P^{(dec)}$ learnable parameters for its encoder and decoder. 
Given that the models are pairwise, for $L$ languages, we have a total of $L(L - 1)$ learnable parameter vectors of size $P$. 
On the other hand, using our contextual parameter generator we have a total of $L$ vectors of size $M$ (one for each language), and a single matrix of size $P \times M$. 
Then, the parameters of the encoder and the decoder, for a single language pair, are defined as a linear combination of the $M$ columns of that matrix.

\paragraph{Controlled Parameter Sharing.} We can further control parameter sharing by observing that the encoder/decoder parameters often have some ``natural grouping''. 
For example, in the case of recurrent neural networks we may have multiple weight matrices, one for each layer, as well as attention-related parameters. 
Based on this observations, we now propose a way to control how much information is shared across languages. 
The language embeddings need to represent all of the language-specific information and thus may need to be large in size. 
However, when computing the parameters of each group, only a small part of that information is relevant. 
Let $\theta^{(enc)} = \smash{\{\theta^{(enc)}_j\}_{j=1}^G}$ and $\theta^{(enc)}_j \in \mathbb{R}^{P^{(enc)}_j}$, where $G$ denotes the number of groups. 
Then, we define:
\begin{align}
	\theta^{(enc)}_j & \triangleq \bf{W^{(enc)}_j} \bf{P^{(enc)}_j} \bf{l}_s,
\end{align}
where $\smash{\bf{W^{(enc)}_j}} \in \mathbb{R}^{P^{(enc)}_j \times M'}$ and $\smash{\bf{P^{(enc)}_j}} \in \mathbb{R}^{M' \times M}$, with $M'<M$ (and similarly for the decoder parameters). 
We can see now that $\smash{\bf{P^{(enc)}_j}}$ is used to extract the relevant information (size $M'$) for parameter group $j$, from the larger language embedding (size $M$).
This allows us to control the parameter sharing across languages in the following way: if we want to increase the number of per-language parameters (i.e., the language embedding size) we can increase $M$ while keeping $M'$ small enough so that the total number of parameters does not explode. 
This would not have been possible without the proposed low-rank approximation for $\bf{W^{(enc)}}$, that uses the parameter grouping information. 

\paragraph{Alternative Options.} Given that our proposed approach does not depend on the specific choice of the parameter generator network, it might be interesting to design models that use side-information about the languages that are being used (such as linguistic information about language families and hierarchies). 
This is outside the scope of this paper, but may be an interesting future direction.

\subsection{Semi-Supervised and Zero-Shot Learning}
\label{sec:semi-sumervised_and_zero-shot_learning}

The proposed parameter generator also enables semi-supervised learning via back-translation. 
Concretely, monolingual data can be used to train the shared encoder/decoder networks to translate a sentence from some language to itself (similar to the idea of auto-encoders by \citet{Vincent:2008}). 
This is possible and can help learning because of the fact that many of the learnable parameters are shared across languages.

Furthermore, zero-shot translation, where the model translates between language pairs for which it has seen no explicit training data, is also possible.
This is because the same per-language parameters are used to translate to and from a given language, irrespective of the language at the other end. 
Therefore, as long as we train our model using some language pairs that involve a given language, it is possible to learn to translate in any direction involving that language.

\subsection{Potential for Adaptation}
\label{sec:domain_adaptation}

Let us assume that we have trained a model using data for some set of languages, $\ell_1, \ell_2, \hdots, \ell_m$. 
If we obtain data for some new language $\ell_n$, we do not have to retrain the whole model from scratch. 
In fact, we can fix the parameters that are shared across all languages and only learn the embedding for the new language (along with the relevant word embeddings if not using a shared vocabulary). 
Assuming that we had a sufficient number of languages in the beginning, this may allow us to obtain reasonable translation performance for the new language, with a minimal amount of training.\footnote{This is due to the small number of parameters that need to be learned in this case. To put this into perspective, in most of our experiments we used language embeddings of size $8$.}


\subsection{Number of Parameters}

For the base case of using multiple pairwise models to perform multilingual translation, each model has $P + 2WV$ parameters, where $P = P^{(enc)} + P^{(dec)}$, $W$ is the word embedding size, and $V$ is the vocabulary size per language (assumed to be the same across languages, without loss of generality). 
Given that the models are pairwise, for $L$ languages, we have a total of $L(L - 1)(P + 2WV) = \mathcal{O}(L^2P +2L^2WV)$ learnable parameters. 
For our approach, using the linear parameter generator network presented in Section \ref{sec:parameter_generator_network}, we have a total of $\mathcal{O}(PM + LWV)$ learnable parameters.
Note that the number of encoder/decoder parameters has no dependence on $L$ now, meaning that our model can easily scale to a large number of languages.



\section{Experiments}


In this section, we describe our experimental setup along with our results and key observations.

\paragraph{Setup.} For all our experiments we use as the base NMT model an encoder-decoder network which uses a bidirectional LSTM for the encoder, and a two-layer LSTM with the attention model of \citet{Bahdanau:2014} for the decoder. 
The word embedding size is set to 512. This is a common baseline model that achieves reasonable performance and we decided to use it as-is, without tuning any of its parameters, as extensive hyperparameter search is outside the scope of this paper.

\begin{table*}[t]
	\caption{Comparison of our proposed approach (shaded rows) with the base pairwise NMT model (PNMT) and the Google multilingual NMT model (GML) for the IWSLT-15 dataset. The \textit{Percent Parallel} row shows what portion of the parallel corpus is used while training; the rest is being used only as monolingual data. Results are shown for the BLEU and Meteor metrics. CPG* represents the same model as CPG, but trained without using auto-encoding training examples. The best score in each case is shown in \textbf{bold}.}
	\label{tab:exp:res:iwslt15}
	\small
	\begin{center}
		\begin{tabu} to \textwidth {|[0.7pt]@{}>{\columncolor{white}[0pt][0pt]}X[1c]@{}|[0.7pt]@{}>{\columncolor{white}[0pt][0pt]}X[2c]@{}|[0.7pt]@{}>{\columncolor{white}[0pt][0pt]}X[2c]@{}|@{}>{\columncolor{white}[0pt][0pt]}X[2c]@{}|@{}>{\columncolor{pale_gray}[0pt][0pt]}X[2c]@{}|@{}>{\columncolor{pale_gray}[0pt][0pt]}X[2c]@{}|[0.7pt]@{}>{\columncolor{white}[0pt][0pt]}X[2c]@{}|@{}>{\columncolor{white}[0pt][0pt]}X[2c]@{}|@{}>{\columncolor{pale_gray}[0pt][0pt]}X[2c]@{}|@{}>{\columncolor{pale_gray}[0pt][0pt]}X[2c]@{}|[0.7pt]}
			\tabucline[0.7pt]{3-10}
			\multicolumn{2}{c|[0.7pt]}{} & \multicolumn{4}{c|[0.7pt]}{\textbf{BLEU}} & \multicolumn{4}{c|[0.7pt]}{\textbf{Meteor}} \\ \tabucline[0.4pt]{3-10}
			\multicolumn{2}{c|[0.7pt]}{} & PNMT  & GML   & CPG*  & CPG            & PNMT  & GML   & CPG*  & CPG \\ \tabucline[0.7pt]{1-10}
			& \texttt{En}$\tr$\texttt{Cs} & 14.89 & 15.92 & 16.88 & \textbf{17.22} & 19.72 & 20.93 & 21.51 & \textbf{21.72} \\ 
			& \texttt{Cs}$\tr$\texttt{En} & 24.43 & 25.25 & 26.44 & \textbf{27.37} & 27.29 & 27.46 & 28.16 & \textbf{28.52} \\
			& \texttt{En}$\tr$\texttt{De} & 25.99 & 15.92 & 26.41 & \textbf{26.77} & 44.72 & 42.97 & 45.97 & \textbf{46.30} \\ 
			& \texttt{De}$\tr$\texttt{En} & 30.93 & 29.60 & 31.24 & \textbf{31.77} & 30.73 & 29.90 & 30.95 & \textbf{31.13} \\ 
			& \texttt{En}$\tr$\texttt{Fr} & 38.25 & 34.40 & 38.10 & \textbf{38.32} & 57.43 & 53.86 & 57.42 & \textbf{57.68} \\ 
			& \texttt{Fr}$\tr$\texttt{En} & 37.40 & 35.14 & 37.11 & \textbf{37.89} & 34.83 & 33.14 & 34.34 & \textbf{34.89} \\ 
			& \texttt{En}$\tr$\texttt{Th} & 23.62 & 22.22 & 26.03 & \textbf{26.33} &   -   &   -   &   -   &   -            \\
			& \texttt{Th}$\tr$\texttt{En} & 15.54 & 14.03 & 16.54 & \textbf{26.77} & 21.58 & 21.02 & 22.78 & \textbf{23.05} \\ 
			& \texttt{En}$\tr$\texttt{Vi} & 27.47 & 25.54 & 28.33 & \textbf{29.03} &   -   &   -   &   -   &   -            \\
			& \texttt{Vi}$\tr$\texttt{En} & 24.03 & 23.19 & 25.91 & \textbf{26.38} & 27.59 & 26.96 & 28.23 & \textbf{28.79} \\ \tabucline[0.7pt]{2-10}
			\multirow{-10}{*}{\rotatebox[origin=c]{90}{100\% Parallel Data}} & \CM\textbf{Mean} & \CM26.26 & \CM24.12 & \CM27.30 & \CM\textbf{27.80} & \CM32.98 & \CM32.03 & \CM33.67 & \CM\textbf{34.01} \\ \tabucline[0.7pt]{1-10}
			& \texttt{En}$\tr$\texttt{Cs} & \p5.71 & \p8.18 & \p8.40 & \textbf{\p9.49} & 12.18 & 14.97 & 15.25 & \textbf{15.90} \\
			& \texttt{Cs}$\tr$\texttt{En} & \p6.64 & 14.56 & 14.81  & \textbf{15.38}  & 13.02 & 20.04 & 19.98 & \textbf{20.87} \\ 
			& \texttt{En}$\tr$\texttt{De} & 11.70  & 14.60 & 15.09  & \textbf{16.03}  & 29.98 & 33.74 & 34.88 & \textbf{36.19} \\ 
			& \texttt{De}$\tr$\texttt{En} & 18.10  & 19.02 & 19.77  & \textbf{20.25}  & 22.57 & 23.27 & 23.65 & \textbf{24.40} \\ 
			& \texttt{En}$\tr$\texttt{Fr} & 24.47  & 25.15 & 24.00  & \textbf{25.79}  & 44.10 & 44.84 & 44.95 & \textbf{46.22} \\ 
			& \texttt{Fr}$\tr$\texttt{En} & 23.79  & 25.02 & 24.55  & \textbf{27.12}  & 26.28 & 26.61 & 26.20 & \textbf{28.18} \\ 
			& \texttt{En}$\tr$\texttt{Th} & \p7.86 & 15.58 & \textbf{18.41} & 17.65   &   -   &   -   &   -   &   -            \\ 
			& \texttt{Th}$\tr$\texttt{En} & \p7.13 & \p9.11 & \textbf{10.19} & 10.14   & 13.91 & 16.32 & 16.78 & \textbf{16.92} \\ 
			& \texttt{En}$\tr$\texttt{Vi} & 18.01  & 17.51 & \textbf{18.92} & 18.90   &   -   &   -   &   -   &   -            \\
			& \texttt{Vi}$\tr$\texttt{En} & \p6.69 & 16.00 & 16.28  & \textbf{16.86}  & 13.39 & 21.01 & 21.34 & \textbf{22.28} \\
 \tabucline[0.7pt]{2-10}
			\multirow{-11}{*}{\rotatebox[origin=c]{90}{10\% Parallel Data}} & \CM\textbf{Mean} & \CM13.01 & \CM16.47 & \CM17.04 & \CM\textbf{17.76} & \CM21.93 & \CM25.10 & \CM25.38 & \CM\textbf{26.37} \\ \tabucline[0.7pt]{1-10}
			& \texttt{En}$\tr$\texttt{Cs} & \p0.49 & \p1.25 & \p1.57 & \textbf{\p2.38} & \p4.60 & \p6.24 & \p6.28 & \textbf{\p8.38} \\
			& \texttt{Cs}$\tr$\texttt{En} & \p1.10 & \p 1.76 & \p1.87 & \textbf{\p4.60} & \p6.29 & \p7.13 & \p7.08 & \textbf{11.15} \\
			& \texttt{En}$\tr$\texttt{De} & \p1.22 & \p4.13 & \p4.06 & \textbf{\p6.46} & 12.23  & 18.29 & 17.61  & \textbf{23.83} \\
			& \texttt{De}$\tr$\texttt{En} & \p1.46 & \p3.42 & \p3.86 & \textbf{\p7.49} & \p7.58 & \p8.79 & \p8.95 & \textbf{13.73} \\
			& \texttt{En}$\tr$\texttt{Fr} & \p2.88 & \p7.74 & \p7.41 & \textbf{12.45}  & 13.88  & 21.29 & 21.80  & \textbf{30.36} \\
			& \texttt{Fr}$\tr$\texttt{En} & \p4.05 & \p5.22 & \p5.06 & \textbf{11.39}  & \p9.58 & \p9.86 & \p9.83 & \textbf{16.34} \\ 
			& \texttt{En}$\tr$\texttt{Th} & \p1.22 & \p5.72 & \p8.01 & \textbf{\p9.26} &   -    &   -   &   -    &   -            \\ 
			& \texttt{Th}$\tr$\texttt{En} & \p1.42 & \p1.66 & \p1.65 & \textbf{\p3.37} & \p6.08 & \p7.22 & \p5.89 & \textbf{\p8.74} \\ 
			& \texttt{En}$\tr$\texttt{Vi} & \p5.35 & \p5.61 & \p5.48 & \textbf{\p8.00} &   -    &   -   &   -    &   -            \\ 
			& \texttt{Vi}$\tr$\texttt{En} & \p2.01 & \p3.57 & \p3.64 & \textbf{\p6.43} & \p7.86 & \p8.76 & \p8.48 & \textbf{12.04} \\ \tabucline[0.7pt]{2-10}
			\multirow{-11}{*}{\rotatebox[origin=c]{90}{1\% Parallel Data}} & \CM\textbf{Mean} & \CM\p2.12 & \CM\p4.01 & \CM\p4.26 & \CM\textbf{\p7.18} & \CM\p8.51 & \CM10.95 & \CM10.74 & \CM\textbf{15.58} \\ \tabucline[0.7pt]{1-10}
		\end{tabu}
		\vspace{-1.0em}
	\end{center}
\end{table*}

During training, we use a label smoothing factor of 0.1 \citep{Wu:2016} and the AMSGrad optimizer \cite{Reddi:2018} with its default parameters in TensorFlow, and a batch size of 128 (due to GPU memory constraints).
Optimization was stopped when the validation set BLEU score was maximized.
The order in which language pairs are used while training  was as follows: we always first sample a language pair (uniformly at random), and then sample a batch for that pair (uniformly at random).\footnote{We did not observe any ``forgetting'' effect, because we keep ``re-visiting'' all language pairs throughout training. It would be interesting to explore other sampling schemes, but it is outside the scope of this paper.} 
During inference, we employ beam search with a beam size of 10 and the length normalization scheme of \citep{Wu:2016}.
We want to emphasize that we did not run experiments with other architectures or configurations, and thus this architecture was not chosen because it was favorable to our method, but rather because it was a frequently mentioned baseline in existing literature. 

All experiments were run on a machine with a single Nvidia V100 GPU, and $24$ GBs of system memory. Our most expensive experiment took about 10 hours to complete, which would cost about $\$25$ on a cloud computing service such as Google Cloud or Amazon Web Services, thus making our results reproducible, even by independent researchers.

\begin{table*}[t]
	\caption{Comparison of our proposed approach (shaded rows) with the base pairwise NMT model (PNMT) and the Google multilingual NMT model (GML) for the IWSLT-17 dataset. Results are shown for the BLEU metric only because Meteor does not support \texttt{It}, \texttt{Nl}, and \texttt{Ro}. CPG\textsuperscript{8} represents CPG using language embeddings of size 8. The ``\textsubscript{C4}'' subscript represents the low-rank version of CPG for controlled parameter sharing (see Section \ref{sec:parameter_generator_network}), using rank 4, etc. The best score in each case is shown in \textbf{bold}.}
	\label{tab:exp:res:iwslt17}
	\small
	\begin{center}
		\begin{tabu} to \textwidth {|[0.7pt]@{}>{}X[1c]@{}|[0.7pt]@{}>{\columncolor{white}[0pt][0pt]}X[2c]@{}|[0.7pt]@{}>{\columncolor{white}[0pt][0pt]}X[2c]@{}|@{}>{\columncolor{white}[0pt][0pt]}X[2c]@{}|@{}>{\columncolor{pale_gray}[0pt][0pt]}X[2c]@{}|@{}>{\columncolor{pale_gray}[0pt][0pt]}X[2c]@{}|@{}>{\columncolor{pale_gray}[0pt][0pt]}X[2c]@{}|@{}>{\columncolor{pale_gray}[0pt][0pt]}X[2c]@{}|@{}>{\columncolor{pale_gray}[0pt][0pt]}X[2c]@{}|@{}>{\columncolor{pale_gray}[0pt][0pt]}X[2c]@{}|[0.7pt]}
			\tabucline[0.7pt]{3-10}
			\multicolumn{2}{c|[0.7pt]}{} & \multicolumn{8}{c|[0.7pt]}{\textbf{BLEU}} \\ \tabucline[0.4pt]{3-10}
			\multicolumn{2}{c|[0.7pt]}{} & PNMT  & GML   & $\text{CPG}^{\text{8}}$  & $\text{CPG}^{\text{8}}_{\text{C4}}$ & $\text{CPG}^{\text{8}}_{\text{C2}}$ & $\text{CPG}^{\text{8}}_{\text{C1}}$ & $\text{CPG}^{\text{64}}_{\text{C8}}$ & $\text{CPG}^{\text{512}}_{\text{C8}}$ \\ \tabucline[0.7pt]{1-10}
			\multirow{17}{*}{\rotatebox[origin=c]{90}{Supervised}} & \texttt{De}$\tr$\texttt{En} & 21.78 & 21.25 & \textbf{22.56} & 20.78  & 22.09  & 21.23 & 21.50 & 22.38 \\
			& \texttt{De}$\tr$\texttt{It} & 13.16 & 13.84 & \textbf{14.73} & 14.34  & 14.43  & 13.84 & 14.34 & 14.11 \\ 
			& \texttt{De}$\tr$\texttt{Ro} & 10.85 & 11.95 & 12.24 & 12.37  & \textbf{12.72}  & 10.37 & 11.32 & 11.94 \\ 
			& \texttt{En}$\tr$\texttt{De} & \textbf{19.75} & 17.06 & 19.41 & 19.04  & 18.42  & 17.04 & 17.46 & 19.29 \\ 
			& \texttt{En}$\tr$\texttt{It} & 27.70 & 25.74 & 27.57 & 27.11  & \textbf{28.21}  & 26.26 & 27.26 & 27.48 \\ 
			& \texttt{En}$\tr$\texttt{Nl} & 24.41 & 22.46 & 24.47 & \textbf{25.15}  & 24.64  & 23.94 & 24.48 & 24.50 \\ 
			& \texttt{En}$\tr$\texttt{Ro} & 19.23 & 18.60 & 20.83 & \textbf{20.96}  & 18.69  & 17.23 & 20.20 & 20.86 \\ 
			& \texttt{It}$\tr$\texttt{De} & 14.39 & 12.76 & 14.61 & \textbf{15.06}  & 14.15  & 13.12 & 14.18 & 14.69 \\ 
			& \texttt{It}$\tr$\texttt{En} & 29.84 & 27.96 & \textbf{30.62} & 30.10  & 29.44  & 29.22 & 29.56 & 30.18 \\ 
			& \texttt{It}$\tr$\texttt{Nl} & 16.74 & 16.27 & 17.99 & \textbf{18.11}  & 18.05  & 17.13 & 17.71 & 17.99\\ 
			& \texttt{Nl}$\tr$\texttt{En} & 26.30 & 24.78 & 26.31 & 26.17  & 25.74  & 26.15  & \textbf{26.33} & 26.20 \\ 
			& \texttt{Nl}$\tr$\texttt{It} & 16.03 & 16.10 & 16.81 & \textbf{17.50}  & 17.03  & 16.81 & 16.89 & 17.09 \\ 
			& \texttt{Nl}$\tr$\texttt{Ro} & 12.84 & 12.48 & 14.01 & \textbf{14.44}  & 12.56  & 11.79 & 12.38 & 13.66 \\ 
			& \texttt{Ro}$\tr$\texttt{De} & 12.75 & 12.21 & 13.58 & \textbf{13.66}  & 13.02  & 12.62 & 12.96 & 13.63 \\ 
			& \texttt{Ro}$\tr$\texttt{En} & 24.33 & 22.88 & 23.83 & 23.88  & 24.20  & 23.58  & \textbf{24.65} & 23.57 \\ 
			& \texttt{Ro}$\tr$\texttt{Nl} & 13.70 & 14.11 & 15.34 & \textbf{15.51}  & 15.11  & 14.65 & 15.29 & 15.19 \\ \tabucline[0.7pt]{2-10}
			& \CM\textbf{Mean}            & \CM18.99 & \CM18.15 & \CM19.68 & \CM\textbf{19.75} & \CM19.28 & \CM18.44 & \CM19.16 & \CM19.74 \\ \tabucline[0.7pt]{1-10}
			\multirow{5}{*}{\rotatebox[origin=c]{90}{Zero-Shot}} & \texttt{De}$\tr$\texttt{Nl} & 12.75 & 12.50  & 12.74 & \textbf{12.80} & 11.65 & 12.41 & 12.67 & 12.75 \\
			& \texttt{It}$\tr$\texttt{Ro} & \p9.97 & \p9.57 & 10.57 & 10.17 & 10.42 & \p9.65 & \textbf{10.69} & 10.32 \\ 
			& \texttt{Nl}$\tr$\texttt{De} & 11.32 & 10.47  & 11.52 & 11.20 & 11.28 & 10.89  & \textbf{11.63} & 11.45 \\
			& \texttt{Ro}$\tr$\texttt{It} & 11.69 & 10.82  & 11.51 & 11.40 & 11.66 & 11.42  & \textbf{11.78} & 11.27 \\ \tabucline[0.7pt]{2-10}
			& \CM\textbf{Mean}            & \CM 11.43 & \CM 10.84  & \CM 11.59 & \CM 11.39 & \CM 11.25 & \CM 11.09 & \CM\textbf{11.69} & \CM11.44 \\
			\tabucline[0.7pt]{1-10}
		\end{tabu}
		\vspace{-1.0em}
	\end{center}
\end{table*}

\paragraph{Experimental Settings.} The goal of our experiments is to show how, by using a simple modification of this model, (i) we can achieve significant improvements in performance, while at the same time (ii) being more data and computation efficient, and (iii) enabling support for zero-shot translation. 
To that end, we perform three types of experiments:
\begin{enumerate}[noitemsep,leftmargin=*]
	\item \uline{Supervised:} In this experiment, we use full parallel corpora to train our models. Plain pairwise NMT models (PNMT) are compared to the same models modified to use our proposed decoupled parameter generator. We use two variants: (i) one which does not use auto-encoding of monolingual data while training (CPG*), and (ii) one which does (CPG). Please refer to Section \ref{sec:semi-sumervised_and_zero-shot_learning} for more details.
	\item \uline{Low-Resource:} Similar to the supervised experiments except that we limit the size of the parallel corpora used in training. However, for GML and CPG the full monolingual corpus is used for auto-encoding training.
	\item \uline{Zero-Shot:} In this experiment, our goal is to evaluate how well a model can learn to translate between language pairs that it has not seen while training. For example, a model trained using parallel corpora between English and German, and English and French, will be evaluated in translating from German to French. PNMT can perform zero-shot translation in this setting using pivoting. This means that, in the previous example, we would first translate from German to English and then from English to French (using two pairwise models for a single translation). However, pivoting is prone to error propagation incurred when chaining multiple imperfect translations. The proposed CPG models inherently support zero-shot translation and require no pivoting.
\end{enumerate}
For the experiments using the CPG model without controlled parameter sharing, we use language embeddings of size $8$. 
This is based merely on the fact that this is the largest model size we could fit on one GPU. 
Whenever possible, we compare against PNMT, GML by \citet{Johnson:2017},\footnote{We use our own implementation of \texttt{GML} in order to obtain a fair comparison, in terms of the whole MT pipeline. We have modified it to use the same per-language vocabularies that we use for our approaches, as the proposed shared BPE vocabulary fails to perform well for the considered datasets.} and other state-of-the-art results.


\paragraph{Datasets.} We use the following datasets:
\begin{itemize}[noitemsep,leftmargin=*]
	\item \uline{IWSLT-15:} Used for supervised and low-resource experiments only (this dataset does not support zero-shot learning). We report results for Czech (\texttt{Ch}), English (\texttt{En}), French (\texttt{Fr}), German (\texttt{De}), Thai (\texttt{Th}), and Vietnamese (\texttt{Vi}). This dataset contains \texttildelow90,000-220,000 training sentence pairs (depending on the language pair), \texttildelow500-900 validation pairs, and \texttildelow1,000-1,300 test pairs.
	\item \uline{IWSLT-17:} Used for supervised and zero-shot experiments. We report results for Dutch (\texttt{Nl}), English (\texttt{En}), German (\texttt{De}), Italian (\texttt{It}), and Romanian (\texttt{Ro}). This dataset contains \texttildelow220,000 training sentence pairs (for all language pairs except for the zero-shot ones), \texttildelow900 validation pairs, and \texttildelow1,100 test pairs.
\end{itemize}

\paragraph{Data Preprocessing.} We preprocess our data using a modified version of the Moses tokenizer \cite{Koehn:2007} that correctly handles escaped HTML characters. 
We also perform some Unicode character normalization and cleaning. 
While training, we only consider sentences up to length 50. 
For both datasets, we generate a per-language vocabulary consisting of the most frequently occurring words, while ignoring words that appear less than 5 times in the whole corpus, and capping the vocabulary size to 20,000 words.

\paragraph{Results.} Our results for the IWSLT-15 experiments are shown in Table \ref{tab:exp:res:iwslt15}. 
It is clear that our approach consistently outperforms both the corresponding pairwise model and GML. 
Furthermore, its advantage grows larger in the low-resource setting (up to $5.06$ BLEU score difference, or a $2.4\times$ increase), which is expected due to the extensive parameter sharing in our model. 
For this dataset, there exist some additional published \mbox{state-of-the-art} results not shown in Tables \ref{tab:exp:res:iwslt15} and \ref{tab:exp:res:iwslt17}. 
\citet{Huang:2018} report a BLEU score of 28.07 for the \texttt{En$\tr$Vi} task, while our model is able to achieve a score of \textbf{29.03}. 
Furthermore, \citet{Ha:2016} report a BLEU score of 25.87 for the \texttt{En$\tr$De} task, while our model is able to achieve a score of \textbf{26.77}.\footnote{We were unable to find reported state-of-the-art results for the rest of the language pairs.} 
Our results for the IWSLT-17 experiments are shown in Table \ref{tab:exp:res:iwslt17}.\footnote{Note that, our results for IWSLT-17 are not comparable to those of the official challenge report \citep{Cettolo:2018}, as we use less training data, a smaller baseline model, and our evaluation pipeline potentially differs. However, the numbers presented for all methods in this paper are comparable, as they were all obtained using the same baseline model and evaluation pipeline.} 
Again, our method consistently outperforms both PNMT and GML, in both the supervised and the zero-shot settings. 
Furthermore, the results indicate that our model performance is robust to different sizes of the language embeddings and the choice of $M'$ for controllable parameter sharing. 
It only underperforms in the degenerate case where $M'=1$. 
It is also worth noting that, in the fully supervised setting, GML, the current state-of-the-art in the multilingual setting, underperforms the pairwise models.

The presented results provide evidence that our proposed approach is able to significantly improve performance, without requiring extensive tuning.

\begin{figure}[t]
	\centering
    \includegraphics[width=1.0\columnwidth,clip,trim=28mm 10mm 30mm 0mm]{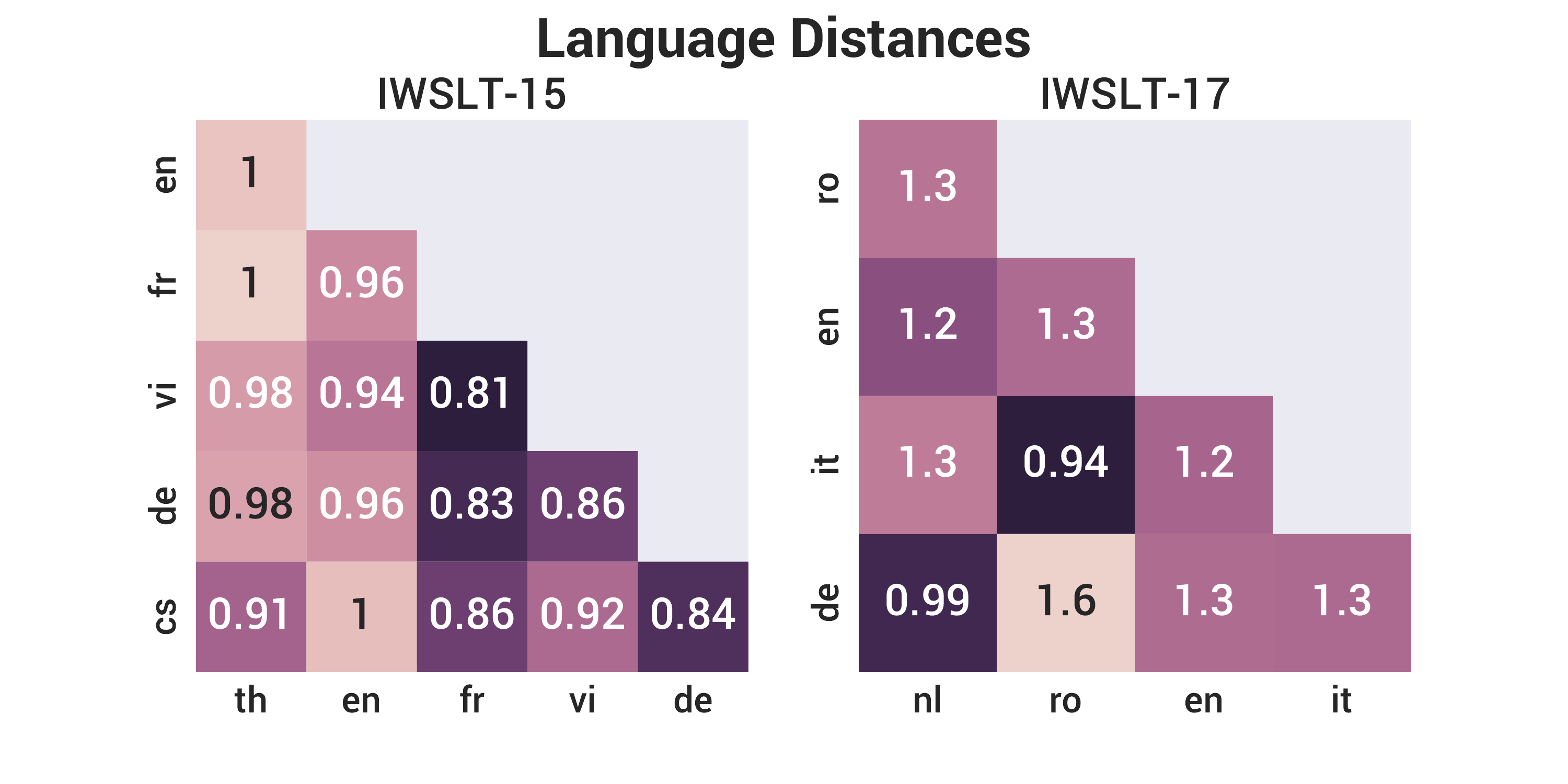}
    \caption{Pairwise cosine distance for all language pairs in the IWSLT-15 and IWSLT-17 datasets. \textbf{Darker} colors represent more similar languages.}
    \vspace{-1.0em}
    \label{fig:language_distances}
\end{figure}

\paragraph{Language Embeddings.} An important aspect of our model is that it learns language embeddings. 
In Figure \ref{fig:language_distances} we show pairwise cosine distances between the learned language embeddings for our fully supervised experiments. 
There are some interesting patterns that indicate that the learned language embeddings are reasonable. 
For example, we observe that German (\texttt{De}) and Dutch (\texttt{Nl}) are most similar for the IWSLT-17 dataset, with Italian (\texttt{It}) and Romanian (\texttt{Ro}) coming second. 
Furthermore, Romanian and German are the furthest apart for that dataset. 
These relationships agree with linguistic knowledge about these languages and the families they belong to.
We see similar patterns in the IWSLT-15 results but we focus on IWSLT-17 here, because it is a larger, better quality, dataset with more supervised language pairs. 
These results are encouraging for analyzing such embeddings to discover relationships between languages that were previously unknown. 
For example, perhaps surprisingly, French (\texttt{Fr}) and Vietnamese (\texttt{Vi}) appear to be significantly related for the IWSLT-15 dataset results. 
This is likely due to French influence in Vietnamese because to the occupation of Vietnam by France during the 19\textsuperscript{th} and 20\textsuperscript{th} centuries \cite{Marr:1981:Language}.

\subsection{Implementation and Reproducibility}

Along with this paper we are releasing an implementation of our approach and experiments as part of a new Scala framework for machine translation.\footnote{\scriptsize\url{https://github.com/eaplatanios/symphony-mt}} 
It is built on top of TensorFlow Scala \cite{Platanios:2018} and follows the modular NMT design (described in Section \ref{sec:nmt_modules}) that supports various NMT models, including our baselines (e.g., \citet{Johnson:2017}). 
It also contains data loading and preprocessing pipelines that support multiple datasets and languages, and is more efficient than other packages (e.g., \texttt{tf-nmt}\footnote{\scriptsize\url{https://github.com/tensorflow/nmt}}). 
Furthermore, the framework supports various vocabularies, among which we provide a new implementation for the byte-pair encoding (BPE) algorithm \cite{Sennrich:2016a} that is 2 to 3 orders of magnitude faster than the released one.\footnote{\scriptsize\url{https://github.com/rsennrich/subword-nmt}}
All experiments presented in this paper were performed using version \texttt{0.1.0} of the framework.

\section{Related Work}

Interlingual translation \cite{Richens:1958} has been the object of many research efforts. 
For a long time, before the move to NMT, most practical machine translation systems only focused on individual language pairs. 
Since the success of end-to-end NMT approaches such as the encoder-decoder framework \cite{Sutskever:2014,Bahdanau:2014,Cho:2014}, recent work has tried to extend the framework to multi-lingual translation. 
An early approach was \citet{Dong:2015} who performed one-to-many translation with a separate attention mechanism for each decoder. 
\citet{Luong:2015a} extended this idea with a focus on multi-task learning and multiple encoders and decoders, operating in a single shared vector space. 
The same architecture is used in \cite{Caglayan:2016} for translation across multiple modalities. 
\citet{Zoph:2016a} flipped this idea with a many-to-one translation model, however requiring the presence of a multi-way parallel corpus between all the languages, which is difficult to obtain. 
\citet{Lee:2016} used a single character-level encoder across multiple languages by training a model on a many-to-one translation task. 
Closest to our work are more recent approaches, already described in Section \ref{sec:background} \cite{Firat:2016a,Johnson:2017,Ha:2016}, that attempt to enforce different kinds of parameter sharing across languages.

Parameter sharing in multilingual NMT naturally enables semi-supervised and zero-shot learning. 
Unsupervised learning has been previously explored with key ideas such as back-translation \cite{Sennrich:2016b}, dual learning \cite{He:2016}, common latent space learning \cite{Lample:2018}, etc. 
In the vein of multilingual NMT, \citet{Artetxe:2017} proposed a model that uses a shared encoder and multiple decoders with a focus on unsupervised translation. 
The entire system uses cross-lingual embeddings and is trained to reconstruct its input using only monolingual data. 
Zero-shot translation was first attempted in \cite{Firat:2016b} who performed zero-zhot translation using their pre-trained multi-way multilingual model, fine-tuning it with pseudo-parallel data generated by the model itself. 
This was recently extended using a teacher-student framework \cite{Chen:2017}. 
Later, zero-shot translation without any additional steps was attempted in \cite{Johnson:2017} using their shared encoder-decoder network. 
An iterative training procedure that leverages the duality of translations directly generated by the system for zero-shot learning was proposed by \citet{Lakew:2017}. 
For extremely low resource languages, \newcite{Gu:2018} proposed sharing lexical and sentence-level representations across multiple source languages with a single target language. 
Closely related is the work of \citet{Cheng:2016} who proposed the joint training of source-to-pivot and pivot-to-target NMT models.

\citet{Ha:2018} are probably the first to introduce a similar idea to that of having one network (called a {\em hypernetwork}) generate the parameters of another. 
However, in that work, the input to the hypernetwork are structural features of the original network (e.g., layer size and index). 
\citet{AlShedivat:2017} also propose a related method where a neural network generates the parameters of a linear model. 
Their focus is mostly on interpretability (i.e., knowing which features the network considers important). However, to our knowledge, there is no previous work which proposes having a network generate the parameters of another deep neural network (e.g., a recurrent neural network), using some well-defined context based on the input data. 
This context, in our case, is the language of the input sentences to the translation model, along with the target translation language.

\section{Conclusion and Future Directions}

We have presented here a novel {\em contextual parameter generation} approach to neural machine translation. 
Our resulting system, which outperforms other state-of-the-art systems, uses a standard pairwise encoder-decoder architecture. 
However, it differs from earlier approaches by incorporating a component that generates the parameters to be used by the encoder and the decoder for the current sentence, based on the source and target languages, respectively.
We refer to this novel component as the {\em contextual parameter generator}.
The benefit of this approach is that it dramatically improves the ratio of the number of parameters to be learned, to the number of training examples available, by leveraging shared structure across different languages. 
Thus, our approach does not require any extra machinery such as back-translation, dual learning, pivoting, or multilingual word embeddings. 
It rather relies on the simple idea of {\em treating language as a context within which to encode/decode}. 
We also showed that the proposed approach is able to achieve state-of-the-art performance without requiring any tuning. 
Finally, we performed a basic analysis of the learned language embeddings, which showed that cosine distances between the learned language embeddings reflect well known similarities among language pairs such as German and Dutch.

In the future, we want to extend the concept of the contextual parameter generator to more general settings, such as translating between different modalities of data (e.g., image captioning). 
Furthermore, based on the discussion of Section \ref{sec:domain_adaptation}, we hope to develop an adaptable, never-ending learning \cite{Mitchell:2018} NMT system.

\section*{Acknowledgments}

We would like to thank Otilia Stretcu, Abulhair Saparov, and Maruan Al-Shedivat for the useful feedback they provided in early versions of this paper. This research was supported in part by AFOSR under grant FA95501710218.

\bibliography{emnlp2018}
\bibliographystyle{acl_natbib}

\appendix

\end{document}